# Design of a Smart Unmanned Ground Vehicle for Hazardous Environments

Saurav Chakraborty

*Tyfone Communications Development (I) Pvt. Ltd.
ITPL, White Field,
Bangalore 560092, INDIA.*

Subhadip Basu

*Computer Science & Engineering. Dept.
Jadavpur University
Kolkata – 700032, INDIA*

**Abstract.** *A smart Unmanned Ground Vehicle (UGV) is designed and developed for some application specific missions to operate predominantly in hazardous environments. In our work, we have developed a small and lightweight vehicle to operate in general cross-country terrains in or without daylight. The UGV can send visual feedbacks to the operator at a remote location. Onboard infrared sensors can detect the obstacles around the UGV and sends signals to the operator.*

## 1. Introduction

Robotics is an important field of interest in modern age of automation. Unlike human being a computer controlled robot can work with speed and accuracy without feeling exhausted. A robot can also perform preassigned tasks in a hazardous environment, reducing the risks and threats of human beings. As for example, a *mobile robot* can rapidly detect, identify, locate and neutralize variety of threats including enemy force activity, chemical and biological agents and impossible terrain or unusable routes or roads.

In this paper, we have discussed the key designed issues and the development process of a smart UGV. In the broad *dictionary* sense, an unmanned ground vehicle is any piece of mechanized equipment that moves across the surface of the ground and serves as a means of carrying or transporting something, but explicitly does not carry human being. The academic community usually refers to UGVs with significant autonomous capabilities as mobile robots. Every possible UGV system needs some organizing principle, based on the characteristics of each system such as:

- ➢ The purpose of the development effort (often the performance of some application-specific mission);
- ➢ The specific reasons for choosing a UGV solution for the application (e.g., hazardous environment, strength or endurance requirements, size limitation etc.);
- ➢ the technological challenges, in terms of functionality, performance, or cost, posed by the application;
- ➢ The system's intended operating area (e.g., indoor environments, anywhere indoors, outdoors on roads, general cross-country terrain, the deep seafloor, etc.);
- ➢ the vehicle's mode of locomotion (e.g., wheels, tracks, or legs);
- ➢ How the vehicle's path is determined (i.e., control and navigation techniques employed).

Within this context, a vehicle system can be *teleoperated* in which navigational guidance is transmitted to the vehicle from an externally situated human operator. An *autonomous* vehicle is one that determines its own course using onboard sensor and processing resources. A ground vehicle can be *semi-autonomous* when the navigation control scheme combines inputs from both an external human operator and onboard sensors to determine the path.

Variety of research work has already been done to develop effective navigation systems for unmanned ground vehicle [1], [2], [3], [4], [5]. The first major mobile robot developed efforts was Shakey, developed in the late 1960s, at Stanford Research Institute. In early 1980s Defence Advanced Research Projects Agency (DARPA) had developed Autonomous Land Vehicle (ALV) to serve as the platform for several application projects for army, navy and air defence systems. The use of unmanned robotic spacecraft drastically reduces the cost of manned space travel and provides high reliability in all safety-critical subsystems. Intelligent Vehicle or Highway Systems (IVHS) are the final thread of UGV development in recent years. The basic objective of these systems is to significantly improve the safety and efficiency of the surface transportation system leading ultimately to complete automated vehicle control technologies.

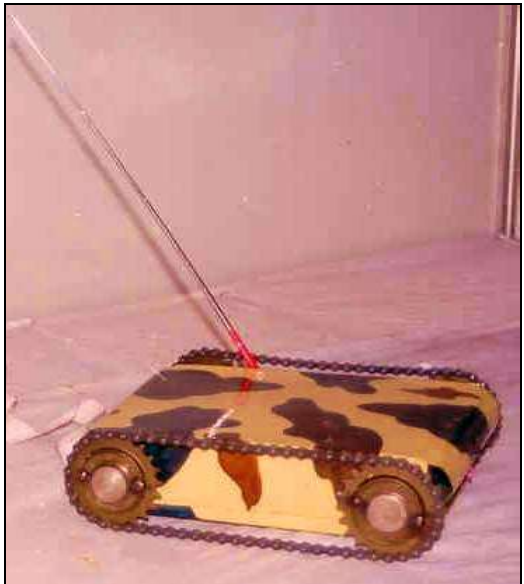
Fig. 1. An image of the UGV

In the current work, we have developed a low cost smart unmanned ground vehicle that can address some of the potential challenges mentioned above. The vehicle navigation guidance system combines inputs from both a remote human operator and onboard sensor systems. The four wheel chained drive mechanism of the vehicle enables easy navigation through rough terrains. Small size and lightweight of the vehicle makes it handy equipment for de-mining operations, bomb disposal and for combat tasks at enemy hideouts (shown in Fig. 1). The UGV can provide real-time intelligence to the remote operator, thus providing them with additional time and distance to effectively make decisions and execute plans, reducing risk and neutralizing threats. The onboard night vision equipment enables the operator to sneak in the enemy hideout during the darkness of night. Effective and efficient designs are the major challenges for the development of an unmanned ground vehicle. The key aspects include the mechanical design of the vehicle, the navigation control system and the onboard sensor systems.

## 2. The Mechanical Design of the Vehicle

The systems intended operation area and surrounding environments govern the mechanical design issues of an UGV. In this work we have assumed a general operational environment with rough terrains and narrow escape routes with minimum space for directional movements. The mechanical structure of the car is shown in the Fig. 2.

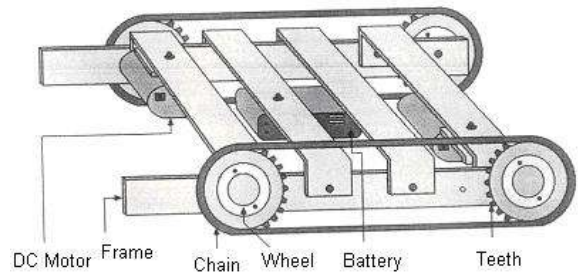
Fig. 2. Mechanical structure of the UGV

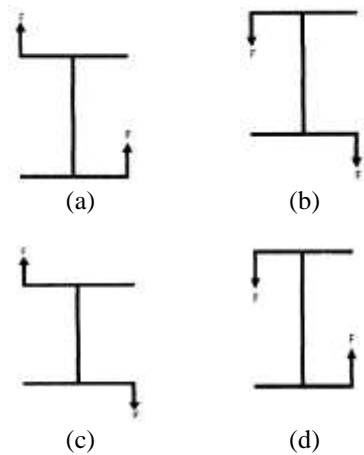
Fig. 3. Free body diagrams of forward (a), backward (b), right (c) and left (d) directional movements for the UGV.

We have designed a chained wheel mechanism where 2 wheels are connected by a thick iron chain. There are 2 chained wheel mechanisms on either side of the vehicle. All the 4 wheels are connected to the vehicles chassis by individual steel shaft and steel ball bearings. Two 12 Volt, 2 A (Maximum load current) D.C. motors are used to drive the pair of chained wheels. The motors are connected to the wheel shaft by 1:10 reduced gear ratio to reduce the speed and to increase the torque of each motor. Each motor can rotate clockwise or anticlockwise direction. The chassis of the vehicle is made up of aluminum strips to reduce the overall weight of the vehicle. The two motors are connected to two diagonally opposite wheels and two other wheels remain free. The diagonally placed motor can provide necessary torque to turn the vehicle at a given angle. Eight relay switches are used to control the motors. Fig 3 (a-d) shows 4 basic directional movements of the vehicle with this driving mechanism. The vehicle can make 180 degree turn with almost zero displacement of its center of gravity.

## 3. Navigation Control System

The most critical challenge for developing an UGV is to design the navigation and guidance systems. In this work, we have installed a wireless analog camera in remote UGV to get visual feedback along with an infrared Search Light for night vision. The distant operator can view the video feedback in the master computer and sends navigation control signals for the UGV. The schematic diagram of the navigation control systems is shown in Fig. 4.

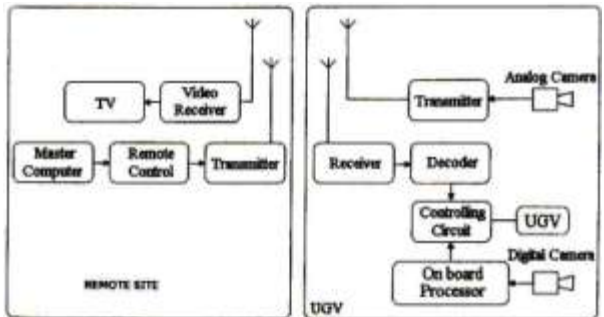

Fig. 4. Schematic diagram of the navigation control system of the UGV

### 3.1 Control mechanism at the operator:

In the operator location, the video receiver receives the analog video signal, sent by the UGV analog camera from remote location. This signal is received by two types of video receiver, one is a 5" BW TV receiver and another is a TV tuner card connected to the master computer. The controlling software is written in Visual Basic. The program can send some basic navigational instructions (left, right, forward, backward, stop) through the parallel port of the master computer to the encoder. The encoder circuit is basically based on Dual Tone Multiple Frequency (DTMF) technique (as shown in Fig. 5). The encoder then sends the proper analog coded signal to the RF transmitter for sending to the remote UGV. The operator on seeing the live video feedback from the remote UGV sends basic navigational instructions (left, right, forward, backward, stop) for correct navigation. During night the operator can switch on the invisible Infra Red search light for navigation in dark without using ordinary searchlight. Use of ordinary searchlight may always attract a risk to be marked.

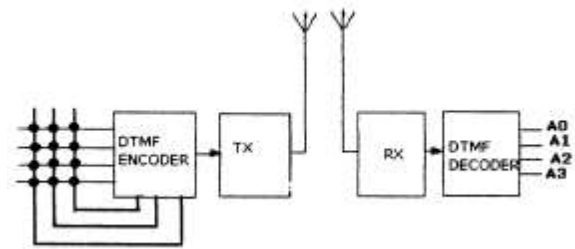

Fig. 5. Schematic of the DTMF encoder and decoder

### 3.2 Control mechanism at the UGV:

The UGV is mounted with fixed analog camera for capturing real-time live video information in front of the vehicle. This serves as an eye of the vehicle and the signal is transmitted to the remote operator through the video transmitter of the UGV. The RF receiver at the UGV receives the coded instruction from the operator. After decoding the instruction appropriate relay switches are switched on or off (as shown in Fig. 6) to complete the instruction using controlling circuit. One of the most important parts of any kind of remotely operated device is its source of power. We have used a 12 Volt, 7 AH sealed lead acid rechargeable battery. With this power source the vehicle can run continuously 4 hours

(approximately). The camera that we have used is sensitive to infra-red (IR) region so we have built a searchlight using 12 IR emitting diodes.

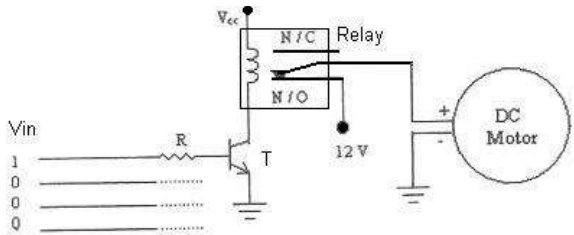

Fig. 6. Block diagram of the Relay circuit

### 4. Onboard Sensor Systems

The infrared movement detector consists of onboard IR transmitter and receiver as shown in Fig. 7. It can detect the obstacle within 2 feet from the sensor. The sensor glows a red LED on detecting an obstacle. The LED is placed below the analog camera to send the visual feedback to the operator.

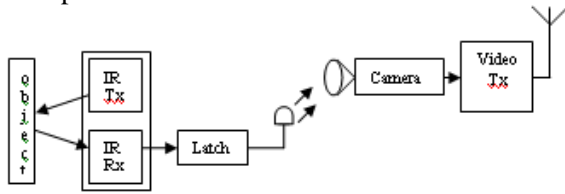

Fig. 7. Schematic of the IR movement detector

### 5. Results and Discussion

In our current work we have successfully developed a prototype unmanned ground vehicle capable of traveling through difficult terrains even without daylight. The major navigational difficulty faced during experimentation is the low ground clearance of the vehicle. Wheels with bigger diameter may solve this problem. The analog camera output is also prone to external noise. Interference from the motor windings often distorts the camera vision at the operator-end.

In one of our earlier works [6], we had developed an online skin colour tracking system with the help of a digital camera, two stepper motors and the stepper motor controlling circuitry. Online recognition of skin colour regions is an important task for many applications such as face recognition, interpretation of gestures, virtual reality interfaces, security monitoring, etc. All have in common the need to track and interpret human activities. The ability to localize and track people's face and hands against a complex background is therefore an important visual problem for the researchers.

To address this need, we are currently developing a smart UGV for tracking of human activities in real-time, based on recognition of skin colours. The single rotating camera tracks the facial and hand movements based on the mean position of the skin colour regions in the image plane captured by the digital camera. The movements of the UGV may be controlled through the visual inputs from the digital camera.

The overall performance of the system may still be improved by inclusion of onboard processing capabilities within the UGV. In our future work, we have planned to develop a lightweight vehicle capable of climbing up and down through stairs.

### Acknowledgement

Authors are thankful to the authorities of MCKV Institute of Engineering for kindly permitting the grant to carry out the research work.